# REAL-TIME VIDEO SUPER-RESOLUTION BY JOINT LOCAL INFERENCE AND GLOBAL PARAMETER ESTIMATION

Noam Elron[1]     Alex Itskovich     Shahar Yuval     Noam Levy

## 1. INTRODUCTION

The state of the art in video super-resolution (SR) are techniques based on deep learning, but they perform poorly on real-world videos (see Figure 1). The reason is that training image-pairs are commonly created by downscaling high-resolution image to produce a low-resolution counterpart. Deep models are therefore trained to *undo downscaling* and do not generalize to super-resolving real-world images. Several recent publications present techniques for improving the generalization of learning-based SR (e.g., [4] and references therein), but are all ill-suited for real-time application.

We present a novel approach to synthesizing training data by simulating two digital-camera image-capture processes at different scales. Our method produces image-pairs in which both images have properties of natural images. Training an SR model using this data leads to far better generalization to real-world images and videos.

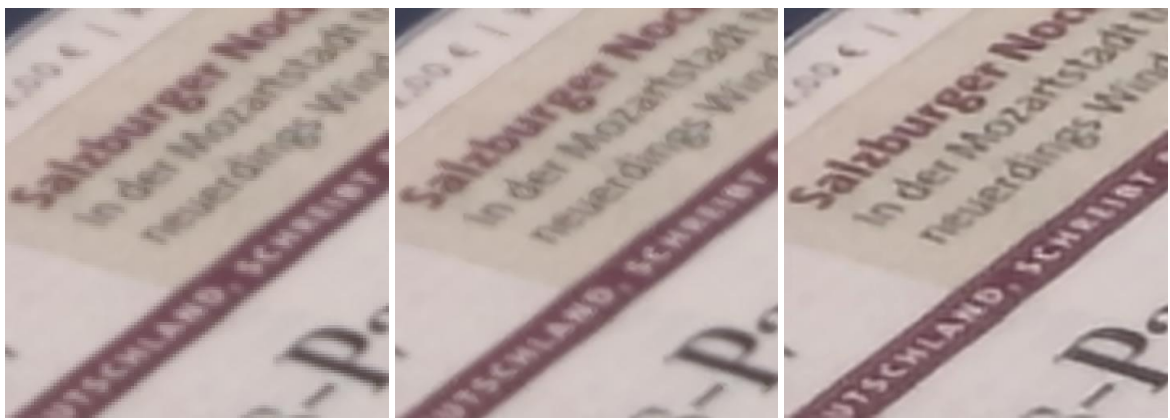

**Original** (displayed at same size as upscaled images)     SR using **DBPN** [3]     SR using our method

*Figure 1. Video SR result for a real-world video (captured using a Sony IMX362 sensor, upscaled ×4). A state-of-the-art method does not improve on the original, whereas our result is substantially sharper.*

In addition, deep video-SR models are characterized by a high operations-per-pixel count, which prohibits their application in real-time. We present an efficient CNN architecture, which enables real-time application of video SR on low-power edge-devices. We split the SR task into two sub-tasks: a *control-flow* which estimates global properties of the input video and adapts the weights and biases of a *processing-CNN* that performs the actual processing. Since the process-CNN is tailored to the statistics of the input, its capacity kept low, while retaining effectivity. Also, since video-statistics evolve slowly, the control-flow operates at a much lower rate than the video frame-rate. This reduces the overall computational load by as much as two orders of magnitude (see Table 1), enabling real-time video SR on edge-devices.

---

[1] noam.elron@intel.com

This framework of decoupling the adaptivity of the algorithm from the pixel processing, can be applied in a large family of real-time video enhancement applications, e.g., video denoising, local tone-mapping, stabilization, etc.

## 2. TRAINING DATA FOR SUPER-RESOLUTION – THE COMMON PRACTICE

Training a learning-based SR model requires image-pairs of the same scene at two resolutions. The common practice is to produce such pairs by downscaling a high-resolution image. The SR system is trained to undo the downscaling (Figure 2). The problem is that the low-resolution images in training have different characteristics than real-world images – the outputs of image downscaling tend to have very sharp features. Thus, the CNN is trained on a specific type of input images and does generalize to other natural images.

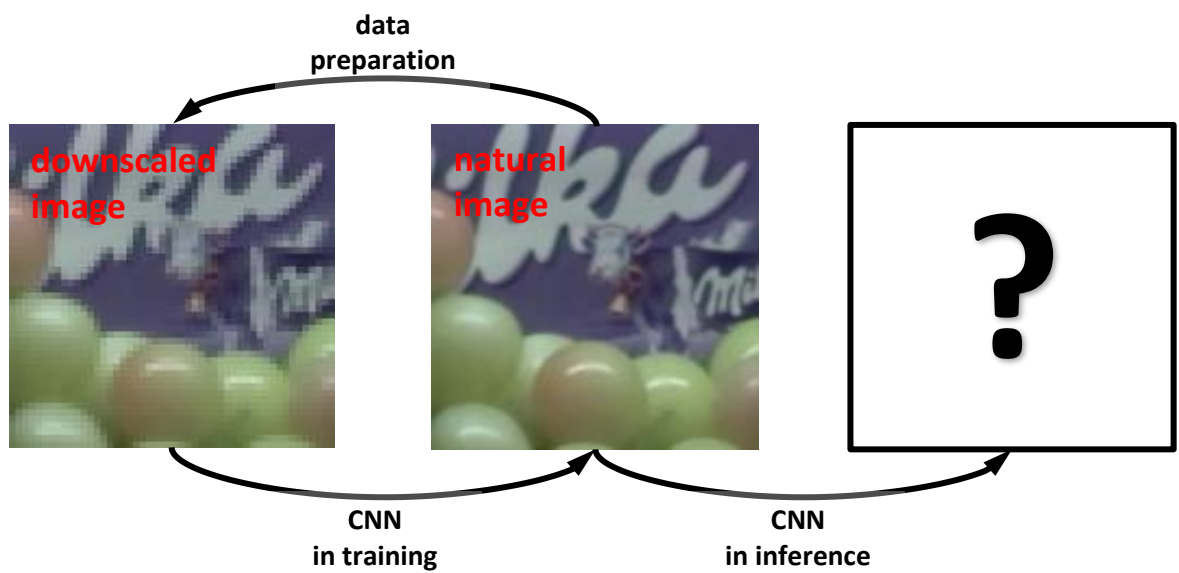

*Figure 2. Training a learning-based SR model – the common practice. The input images during training are the result of a downscaling operation and therefore do not represent real-world images.*

## 3. TRAINING DATA FOR SUPER-RESOLUTION – NOVEL METHOD

Our novel method for creating image pairs for training SR models produces input images with properties that are better correlated with the input images at runtime. Thus learning-based SR systems trained using this data are more likely to generalize to real world videos.

Our method imitates two image-capture processes of the same scene at different scales. For our purpose, the most important transformations that take place when an image is captured using a digital camera are (Figure 3):

(1) Blurring of the scene by the camera lens. The blur is characterized by the lens' *point-spread function* (PSF) – a quantitative measurement of how each infinitesimal point of light in the scene is spread on the sensor plane.

(2) Sampling by the camera sensor.

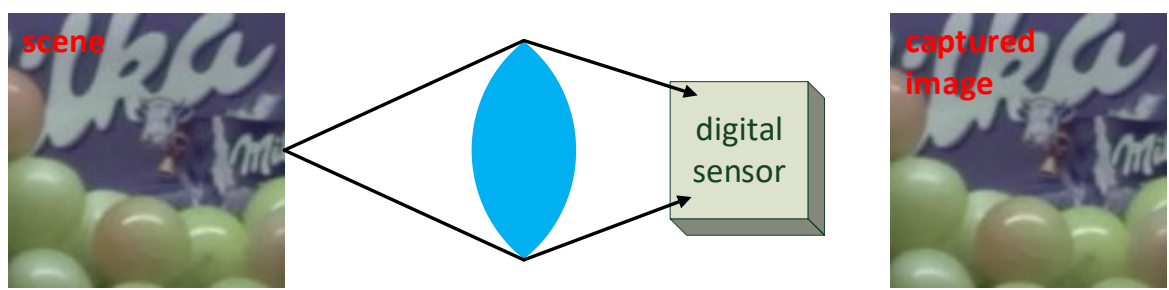

*Figure 3. Image capture process (simplified)*

We use a digital simulation of this simplified image-capture process. Our digital simulation, illustrated in Figure 4, is based on sub-sampling (a.k.a. decimating) a high-resolution image by factor S. This approximates the sampling by the sensor (transformation #2 above). Prior to decimating, we apply the desired lens blur (transformation #1 above). To compensate for the sub-sampling by factor S, applying the blur is done by first spatially stretching the PSF by a factor S and then using convolution.

The resulting image resembles images that have been captured with a lens characterized by the specific PSF. Note that this process has a free parameter S – the larger the factor S, the closer the simulation to actual image-capture of a continuous scene by a digital camera. The resulting image is S times smaller than the input high-resolution image.

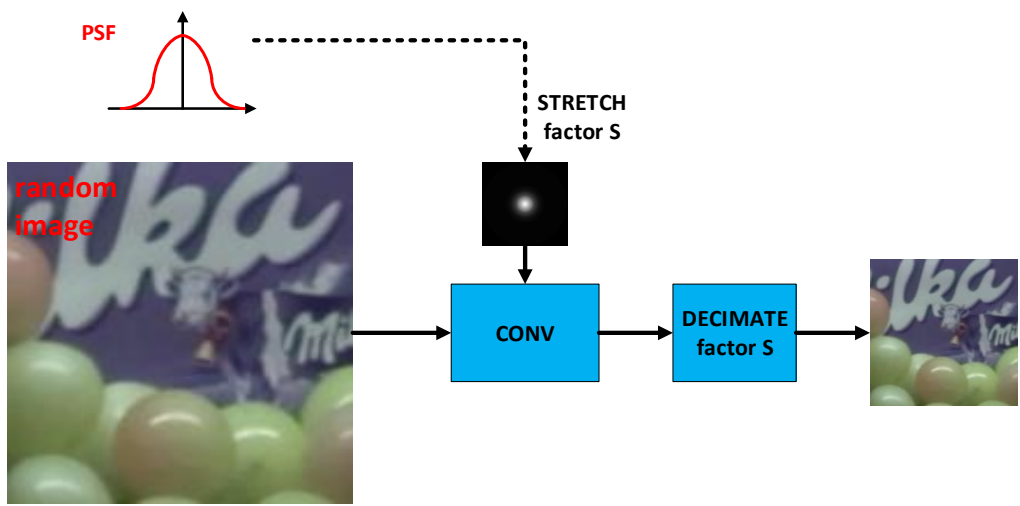

*Figure 4. Digital simulation of image-capture process*

In order to prepare image-pairs for training a learning-based SR model, we perform two such image-capture simulations with factors S and S×R, where R is the upscaling ratio of the SR model (Figure 5). The output is two representations of the scene at different scales with ratio R between them. The smaller of the two images serves as input during training, and the larger as the label – enabling the training of a SR system to upscale by a factor R.

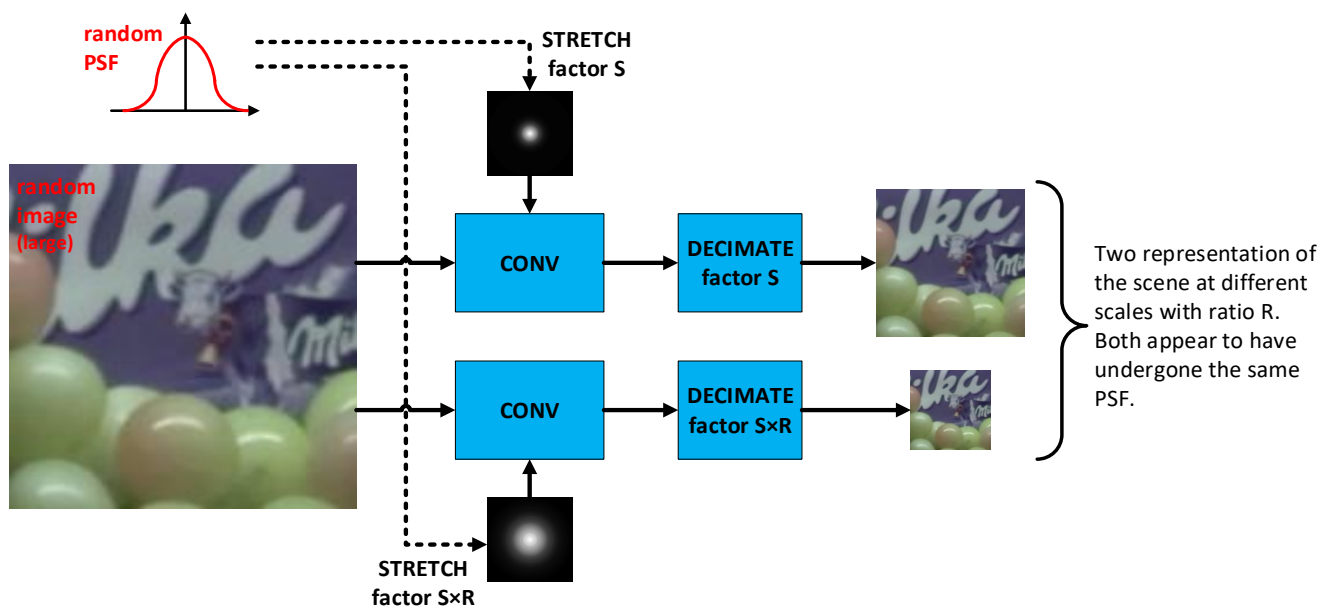

*Figure 5. Novel method of image-pair generation for training of super-resolution models.*

**Choice of PSF**   Training-pairs can be generated with a specific PSF, which would train an SR model to expect images originating from a specific imaging device. Alternatively, a random PSF can be used for each pair, which would train the SR model to expect a variety of real-world images.

**Realistic Image-Capture**   The image-capture simulation described above (and in Figures 3, 4 and 5) is the minimal simulation for our purposes. By adding more transformations that images commonly undergo – e.g., tone-mapping, additive sensor-noise, etc. – the simulation can be made more realistic, increasing the likelihood of generalization.

## 4. VIDEO SUPER-RESOLUTION – HIGH-LEVEL SYSTEM ARCHITECTURE

High-quality video SR is very sensitive to the statistical properties of the input video stream, particularly to the PSF of the input [1, 2]. State-of-the-art SR models (based on deep-learning) are very large, and compute-intensive (see examples in [1, 2, 3]) since they implicitly perform both statistics-estimation and processing.

Our novel architecture decouples the statistics-estimation and the pixel processing. We leverage the fact that in a video stream the statistical properties evolve very slowly and perform the estimation process at a lower rate than the video frame rate, thus reducing the number of operations per pixel by as much as two orders of magnitude (see Table 1).

Figure 6 is a high-level block-diagram of our self-configuring video super-resolution system. It is made up of two distinct flows:

1. The **pixel flow** processes the input pixels to produce super-resolved (upscaled) video frames.
2. The **configuration flow** configures the PROCESS CNN so that the processing is adapted to the current statistical properties of the input video and to the user preferences. The output of the configuration flow are the weights and biases of the PROCESS CNN.

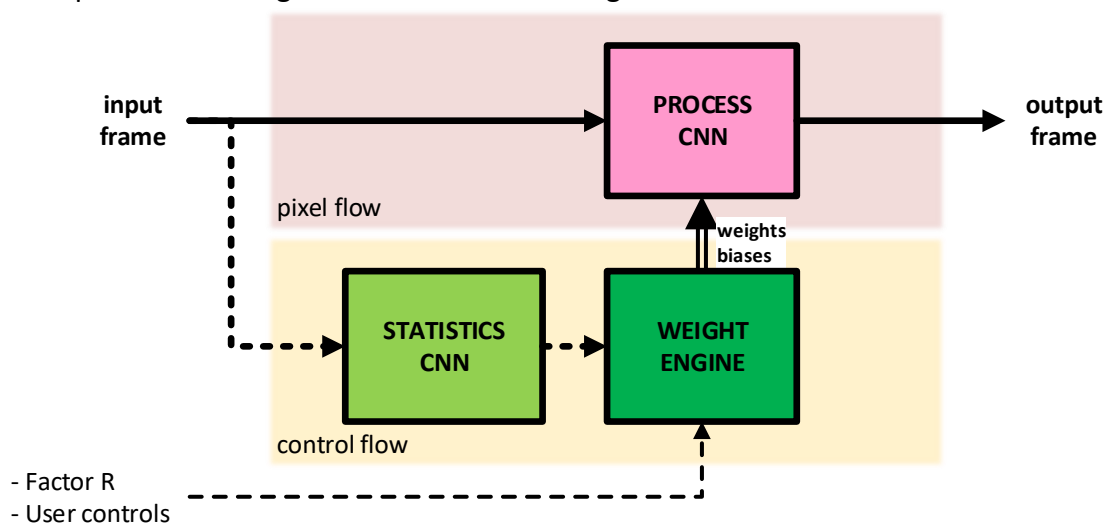

*Figure 6. High-level block diagram of the self-configuring video super-resolution system*

In the configuration flow, the input video frame is fed into a STATISTICS CNN, which outputs an encoding of frame-wide statistics. This information, along with user preferences, is fed into a WEIGHT ENGINE, which produces the weights and biases used in the PROCESS CNN to process the pixels.

The weights and biases used for processing the pixels are highly specialized for the current input statistics and the user preferences. Thus, the PROCESS CNN is focused only on local image features and needs far fewer resources than conventional CNNs used for image/video super-resolution.

Since we can assume with confidence that the statistical properties of the input video stream evolve slowly over time, the configuration flow can be triggered at a lower rate than the frame rate (e.g., every 500-1000 milliseconds). The decoupling of the configuration flow from the frame rate offers several important advantages:

- The configuration flow can be made more powerful without effecting the pixel throughput. If the computation demands increase the trigger rate can be lowered to compensate. This extra power can potentially improve the specialization of the processing weights which may in turn
    - improve the quality of the output video
    - reduce the load on of the PROCESSING CNN, which is the bottleneck of the system.

- The configuration flow can be computed on a separate low power computing device, freeing up resources of the main processing device (e.g., GPU or HW accelerator). The information bandwidth between the two flows is negligible and latency in the computation of the configuration can be tolerated without much effect on the output video (again, due to the slow evolution of the statistics).

- Adjusting visual attributes of the output video (e.g., sharpness) based on user preference is possible with negligible overhead. In common deep-learning systems this desirable property comes at a large price – the neural network must be made more powerful to accommodate extra flexibility. Externalizing this additional degree of adaptivity to the configuration flow eliminates increased load on the PROCESS CNN.

### 5. COMPARISON WITH PUBLISHED STATE-OF-THE-ART SYSTEM

We compare our novel video SR architecture with a representative state-of-the-art SR model. We assume the output video is 1080p at 30 frames-per-second with ×4 upscaling (i.e., the input frame size is 270×320). The numbers given below are for the variants of each method which produced the results in Figure 1.

|  | NDBP [3] | Our architecture (control flow is triggered once per 10 frames) |
|---|---|---|
| # parameters | 10,426,358 | 440,742 in control flow<br>12,360 in pixel flow |
| # ops per output pixel | 345,328 | 9,574 |
| # TOps ($10^{12}$) per second | 14.3 | 0.397 |
| Quality* | Does not generalize to real-world images (see Fig. 1) | State-of-the-art on real-world videos |

*Table 1. Comparison with state-of-the-art SR model. (\*) quality as measured by common practice subjective evaluation [MOS scoring], objective quality cannot be measured when upscaling real-world videos (no 'ground truth' is available).*

### 6. TOPOLOGY OF THE STATISTICS CNN (EXAMPLE)

Figure 7 is an example topology of a STATISTICS CNN. The input image is fed through several convolutional layers (convolution and activation), and then the feature maps undergo *global weighted averaging*. The result of the global averaging is a vector which characterizes the entire video frame. This is an encoding of the statistical properties of the frame and is one of the inputs to the WEIGHT ENGINE (the second input is user preferences).

As stated above, by adjusting the trigger rate of the configuration flow, the STATISTICS CNN can be made arbitrarily large (powerful) without effecting the overall computation costs. This means that very detailed statistical descriptions are potentially possible.

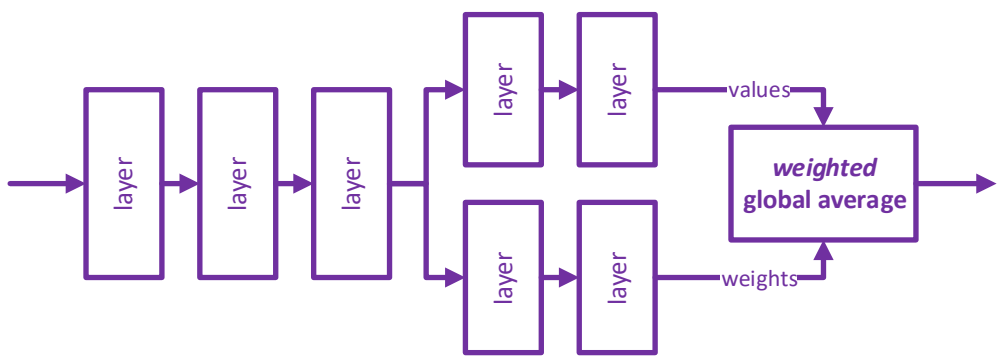

*Figure 7. Example topology of a Statistics-CNN. The global weighted average produces a vector representation of the entire frame.*

### 7. WEIGHT ENGINE (EXAMPLE)

The role of the WEIGHT ENGINE is to supply the convolution kernels (and biases) used by the PROCESS CNN to process the video frames. It receives the frame statistics and user preferences and uses them to supply kernels that are particular for the current setting.

A possible mechanism for producing such an effect is to produce the process-kernel by combining several kernels based on the statistics/preferences. As a naïve example, assume the statistics are encoded as a single number $\alpha$ in the range between 0 and 1. The WEIGHT ENGINE then supplies the combined kernel

$$ker_{\text{process}} = a \cdot ker_1 + (1 - \alpha) \cdot ker_2$$

which depends on the current statistic $a$.

A general WEIGHT ENGINE takes in a more complex statistics/preferences vector. It can be comprised of a large collection of convolution kernels $ker_{ij}$ and a fully connected neural network (FCNN) that controls combinations of the kernels. The statistics vector and the user preferences are the inputs to the FCNN. The output of the FCNN are the coefficients used to linearly combine the convolution kernels.

$$\alpha = \text{FCNN}(\text{statistics vector}, \text{user preference vector})$$

$$ker_j = \sum_i \alpha_{ij} \, ker_{ij}$$

The mixed convolution kernel $ker_j$ is then used as the weights in the $j^{\text{th}}$ convolution of the PROCESSING CNN. A similar mechanism can produce the biases. The convolution kernels $ker_{ij}$ can be trained parameters of the deep-learning system.

### 8. TOPOLOGY OF THE PROCESS CNN (EXAMPLE)

Since the weights and biases of the PROCESS CNN are highly tailored to the current imaging conditions, it can have very low capacity and still perform effectively. An example topology is given in Figure 8. It is a simple feed-forward topology with two characteristics common in topologies for SR – a *residual-connection* which is upscaled using bicubic interpolation, and a *pixel-shuffle* operation wherein a tensor of size $W \times H \times n^2$ is shuffled to produce a single channel of size $n \cdot W \times n \cdot H$.

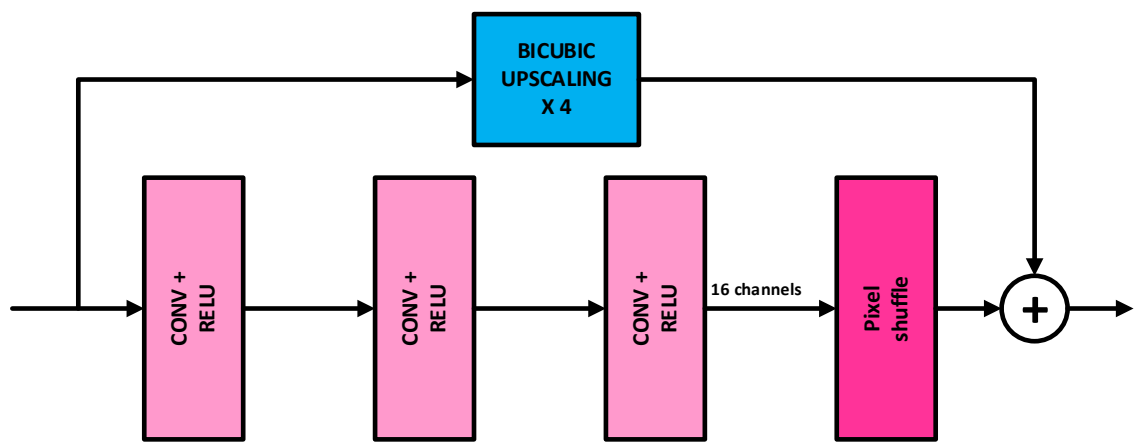

*Figure 8. Example topology of a Process-CNN - a low-capacity feed-forward design*

## 9. TRAINING THE DEEP-LEARNING VIDEO SR SYSTEM

All the operations of the configuration flow and the PROCESSING CNN can be made differentiable with respect to all the parameters. This is the case with the example STATISTICS CNN and example WEIGHT ENGINE described above. Thus, the entire system can be trained end-to-end using common gradient-descent algorithms for training deep-learning systems.

## 10. OTHER REAL-TIME VIDEO PROCESSING TASKS

The framework detailed above of decoupling the adaptivity of the algorithm from the pixel processing, can be applied in a large family of real-time video enhancement applications. For example, this framework may be applicable in video denoising, local tone-mapping, stabilization, etc.

## 11. REFERENCES

[1] W. Yang, X. Zhang, Y. Tian, W. Wang, J. Xue and Q. Liao, *Deep Learning for Single Image Super-Resolution: A Brief Review*, in *IEEE Transactions on Multimedia*, vol. 21, no. 12, 2019.

[2] Z. Wang, J. Chen and S. C. H. Hoi, *Deep Learning for Image Super-resolution: A Survey*, in *IEEE Transactions on Pattern Analysis and Machine Intelligence*, 2020

[3] M. Haris, G. Shakhnarovich and N. Ukita, *Deep Back-Projection Networks for Super-Resolution*, CVPR 2018.

[4] S. Bell-Kligler, A. Shocher and M. Irani, *Blind Super-Resolution Kernel Estimation using an Internal-GAN*, NIPS 2019.